\documentclass{article}

\usepackage[main,preprint]{neurips_2026}

\usepackage[utf8]{inputenc} 
\usepackage[T1]{fontenc}    
\usepackage{hyperref}       
\usepackage{url}            
\usepackage{booktabs}       
\usepackage{amsfonts}       
\usepackage{nicefrac}       
\usepackage{microtype}      
\usepackage{xcolor}         
\usepackage{multirow}
\usepackage{algorithm}
\usepackage{algpseudocode}
\usepackage{amsmath}
\usepackage{booktabs}
\usepackage{graphicx}
\usepackage{makecell}
\usepackage{wrapfig}
\usepackage{arydshln}
\newcommand{\NAME}{
{\textsc{CO-ALIGN}}}
\usepackage{pifont}
\newcommand{\cmark}{\textcolor{green!60!black}{\ding{51}}}
\newcommand{\xmark}{\textcolor{red}{\ding{55}}}
\title{Efficient bias mitigation in T2I diffusion models using Concept Graphs}

\author{%
  Mansi \\
  Department of Computing\\
  Imperial College London\\
  \texttt{m.-24@imperial.ac.uk} \\
  \And
  Avinash Kori \\
  Department of Computing\\
  Imperial College London\\
  \texttt{a.kori21@imperial.ac.uk} \\
  \And
  Francesco Leofante \\
  Department of Computing\\
  Imperial College London\\
  \texttt{f.leofante@imperial.ac.uk} \\
}

\begin{document}

\maketitle

\begin{abstract}
Text-to-Image diffusion models often propagate harmful bias inherited from the training data. Existing bias mitigation techniques typically intervene only at the text encoder or provide inference-time guidance, often leading to generations that collapse into \emph{semantically incoherent outputs}. To address these limitations, we introduce \NAME{} (Concept Ontology Alignment), a novel bias mitigation approach based on concept-graph alignment that operates on the model's \emph{internal concept ontology}. By aligning concepts within the text encoder and denoiser, \NAME{} achieves substantial bias reduction while preserving generative integrity. 
We demonstrate the effectiveness of concept-graph alignment across three paradigms: text-encoders, denoisers and joint text-denoiser ontology alignment. 
\NAME{} outperforms the state of the art, improving fairness by $30\%$, $\Delta FID=11.4$ in image quality, $2.8\%$ in image fidelity, all while reducing semantically incoherent outputs by $88\%$. Beyond bias mitigation, we show that CO-ALIGN benefits other downstream tasks as well. In particular, our experiments demonstrate that better-aligned internal ontologies enhance concept unlearning robustness across multiple unlearning techniques.
\end{abstract}

\section{Introduction}
\label{sec:intro}

Text-to-image (T2I) diffusion models have achieved remarkable generative capability, yet they systematically reproduce and amplify the demographic biases present in their training corpora~\citep{schuhmann2022laion,bianchi2023easily}. These biases are not just superficial stylistic artifacts, but structural properties of the model's learned concept associations. Once embedded in the weights, they are reproduced consistently at every generation. To address this, current bias mitigation methods intervene at one of two levels: the \emph{text encoder}, via text embedding fine-tuning~\citep{shen2024finetuning} or inference-time steering~\citep{friedrich2023fairdiffusion,chuang2023debiasing}; or the \emph{denoiser}, via cross-attention weight editing~\citep{orgad2023time,gandikota2024uce} or latent space manipulation~\citep{parihar2024balancing,li2024selfdiscovering,shi2025difflens}.
Both paradigms share a common and consequential assumption: that the text encoder and denoiser can be treated as \emph{independent} correction targets, and that fixing one is sufficient to correct the generated output. 

\textbf{The problem with existing approaches.}  
The denoiser usually adapts to the given text-encoder, or in the case of end-to-end training, both text encoder and the denoiser are jointly updated, without explicit \emph{disentanglement} guarantees:
the denoiser's cross-attention heads learn to expect the \emph{specific biased embedding geometry} produced by the original text encoder, and have organized their internal concept representations around producing biased visual outputs. When only the text encoder is edited, the denoiser receives an out-of-distribution conditioning signal, which it often cannot interpret coherently, resulting in \emph{semantically incoherent} generations that depict neither the target concept nor any meaningful content.
Conversely, when only the denoiser is edited, the text encoder continues to route underspecified prompts (e.g., \textit{``a nurse''}, where the gender is not mentioned) toward biased attribute embeddings, partially negating the denoiser-side correction.
 \begin{figure}[t]
    \centering
    \includegraphics[width=0.8\linewidth]{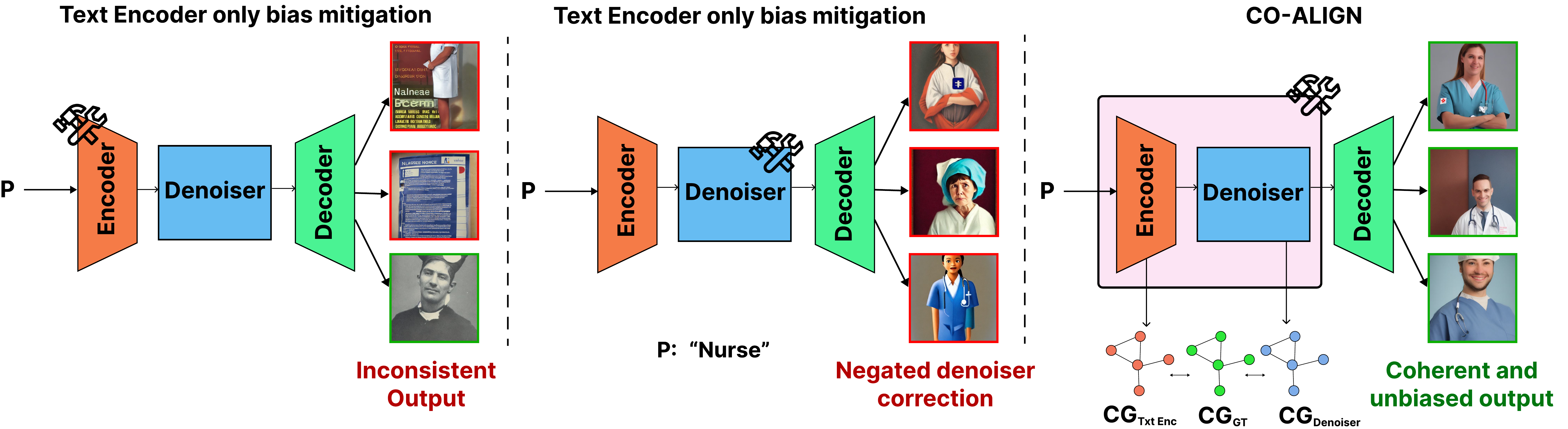}
    \caption{Existing bias mitigation methods that intervene on only the text encoder (left) or denoiser (centre) produce either semantically incoherent outputs or gender-biased generations for the prompt \textit{``Nurse''}. \NAME{} (right) jointly aligns the concept graphs of both components toward a target topology, restoring both coherence and demographic balance. CG representes the concept graphs.}
    \label{fig:placeholder}
    \vspace{-10pt}
\end{figure}

\textbf{Our approach.} We propose to mitigate this by aligning the model's \emph{internal concept ontology} directly. We propose \NAME{} (\textbf{C}oncept  \textbf{O}ntology \textbf{ALIGN}ment), a framework that extracts and edits the structured relational graph of \emph{concept} representations distributed across both the text encoder's embedding space and the denoiser's cross-attention layers. 
The key insight underlying \NAME{} is that bias is not a property of individual concept representations, but of the \emph{relationships between them}. 
In the original model, biased concepts share disproportionately more neural substrate with their stereotypically associated attributes than with their counter-stereotypical ones, like a nurse if majorly attributed to a `female' nurse over a `male' nurse. 
This asymmetry is observable in the extracted concept graph via asymmetric edge weights.
\NAME{} addresses this by aligning the concept graph toward a \emph{target topology}, bringing under-represented concept associations closer to the target concept, jointly in both the text encoder and the denoiser, consequently restoring balanced representation in the final generation. As shown in our experiments, this joint editing enables alignment of the geometric representations in the text encoder and the denoiser, leading to faithful and unbiased generation.

\textbf{Emergent property in \NAME{}.}
Beyond our primary debiasing objective, we find that aligning concept graphs through \NAME{} causes \emph{semantic neighbourhood propagation}, i.e. \emph{aligning one concept pulls its semantic neighbourhood along with it.} This effect is analogous to label propagation in graph-based semi-supervised learning~\citep{zhou2003learning,zhu2003semi,iscen2019label}, where supervision on a small set of labelled nodes propagates to adjacent unlabeled nodes via the graph smoothness assumption~\citep{belkin2006manifold}. 
%
We leverage this property, along with the locality of concept erasure property of unlearning techniques (\cite{bui2025fantastic}), for a second downstream application of robust unlearning. By aligning a small set of adversarial bypass concepts toward the unlearning target, before applying any unlearning technique, \NAME{} significantly improves the robustness of post-hoc unlearning techniques against adversarial prompts.
 
\textbf{Summary of contributions.} Our main contributions can be summarized as follows.
\ding{182} We show that current single-component debiasing techniques trade off bias for inconsistent generation and residual bias. 
\ding{183} Motivated by this analysis, we propose \NAME{}, a bias mitigation framework that extracts a concept-level graph from both the text encoder and the denoiser, and aligns it toward a target topology jointly across both components. 
\ding{184} We demonstrate that \NAME{} outperforms existing state of the art debiasing techniques in producing class balanced and coherent outputs while preserving the generative quality of the output.
\ding{185} We empirically characterize a neighbourhood propagation effect and leverage it for significantly improving the adversarial robustness of existing unlearning techniques using \NAME{}.

\section{Inherent Biases in T2I Diffusion Models}
 
\textbf{Dataset biases and their implications.}\label{subsec:dataset-bias-and-implication} Alike other generative models, biases in T2I diffusion models often originate from the large-scale web-scraped corpora used for training. LAION-5B~\citep{schuhmann2022laion}, used for training Stable Diffusion, mirrors societal inequalities at internet scale across profession~\citep{bianchi2023easily}, race~\citep{cho2023dall,luccioni2023stable} and culture~\citep{naik2023social}. These biases have shown to be \emph{amplified} beyond the degree present in training data both during model training~\citep{seshadri2023bias,perera2023analyzing} and inference ~\citep{roos2026bias}. Large scale deployment of these models in both human operated and automated systems have shown detrimental societal consequences of reinforcement of stereotypes~\citep{bianchi2023easily}, across gender, age, race, and geography simultaneously~\citep{naik2023social}.

\textbf{Bias mitigation techniques.} Existing debiasing techniques can be broadly classified in two groups, depending on where in the generation pipeline they intervene, as summarised in Table~\ref{tab:related_works}.

\textit{Text encoder debiasing} methods intervene solely at the text encoder.
Shen et al.~\citep{shen2024finetuning} fine-tune the text encoder directly using a distributional alignment loss on generated images.
Fair Mapping~\citep{li2025fairmapping} learns a lightweight linear remap over conditioning embeddings to project them into a debiased subspace.
Chuang et al.~\citep{chuang2023debiasing} achieve a similar effect training-free, via a calibrated projection matrix that removes biased directions from text embeddings at inference time.
Fair Diffusion~\citep{friedrich2023fairdiffusion} via SEGA~\citep{brack2023sega}, FairGen~\citep{kang2025fairgen}, and Kim et al.~\citep{kim2025rethinking} take a softer approach, steering the conditioning signal or noise initialisation at inference without any weight modification.
While efficient, all text-encoder-only methods leave the denoiser's internal concept organization and the visual biases encoded. 

\textit{Denoiser debiasing} methods target the denoiser's internal representations.
TIME~\citep{orgad2023time} and UCE~\citep{gandikota2024uce} edit cross-attention key-value projections in closed form to reroute underspecified prompts toward target attribute embeddings.
Asyrp~\citep{kwon2023asyrp} establishes the UNet bottleneck (h-space) as a semantically linear space for controllable editing; building on this, Balancing Act~\citep{parihar2024balancing} trains a lightweight predictor on h-space features, Li et al.~\citep{li2024selfdiscovering} discover fairness-sensitive directions without external classifiers, Vardhana et al.~\citep{vardhana2026selfdebias} steer denoising toward a uniform attribute distribution in an unsupervised manner, and SCALEX~\citep{zeng2026scalex} maps conceptual structure via prompt-aligned latent directions.
DIFFLENS~\citep{shi2025difflens} takes a mechanistic approach, using sparse autoencoders to identify and suppress neuron-level dimensions responsible for bias; BiasMap~\citep{chakraborty2025biasmap}, EFA~\citep{park2025efa}, and Yasser et al.~\citep{yasser2026clip} further reveal that many such corrections reduce distributional gaps without disentangling the underlying concept coupling.
MAS~\citep{zhou2024mas} addresses the related problem of association-engendered stereotypes from co-generation of multiple concepts.
While these methods edit the denoiser, they are applied independently of the text encoder, leaving the biased embedding geometry of underspecified prompts uncorrected.

\begin{table}[ht]
\centering
\footnotesize
\setlength{\tabcolsep}{4pt}
\renewcommand{\arraystretch}{0.9}
\caption{Structured comparison of bias mitigation methods across six dimensions, encompassing training dynamics (columns [1--3]), editing region (columns [4--5]) and scalability (column [6]). CO-ALIGN edits both the text encoder and the denoiser while supporting multi-concept joint editing with no added inference cost.}
\label{tab:related_works}
\resizebox{\columnwidth}{!}{
\begin{tabular}{l|ccc|cc|c}
\toprule
\textbf{Method} &
  \shortstack{\textbf{Weight} \\ \textbf{Mod.}[1]} &
  \shortstack{\textbf{Training} \\ \textbf{Free}[2]} &
  \shortstack{\textbf{Infer.} \\ \textbf{Cost}[3]} &
  \shortstack{\textbf{TE} \\ \textbf{Edit}[4]} &
  \shortstack{\textbf{Den.} \\ \textbf{Edit}[5]} &
  \shortstack{\textbf{Multi-} \\ \textbf{Concept}[6]} \\
\midrule
Fair Diffusion~\cite{friedrich2023fairdiffusion}   & \xmark & \cmark & \cmark & \xmark & \xmark & \xmark \\
Debiasing VLMs~\cite{chuang2023debiasing}          & \xmark & \cmark & \cmark & \cmark & \xmark & \xmark \\
FairGen~\cite{kang2025fairgen}                     & \xmark & \cmark & \cmark & \xmark & \xmark & \xmark \\
Kim et al.~\cite{kim2025rethinking}                & \xmark & \cmark & \cmark & \xmark & \xmark & \xmark \\
EFA~\cite{park2025efa}                             & \xmark & \cmark & \cmark & \xmark & \cmark & \xmark \\
H-Distribution~\cite{parihar2024balancing}         & \xmark & \cmark & \cmark & \xmark & \cmark & \xmark \\
Latent Direction~\cite{li2024selfdiscovering}      & \xmark & \cmark & \cmark & \xmark & \cmark & \xmark \\
SelfDebias~\cite{vardhana2026selfdebias}           & \xmark & \cmark & \cmark & \xmark & \cmark & \xmark \\
SCALEX~\cite{zeng2026scalex}                       & \xmark & \cmark & \cmark & \xmark & \cmark & \xmark \\
DiffLens~\cite{shi2025difflens}                    & \xmark & \cmark & \cmark & \xmark & \cmark & \cmark \\
BiasMap~\cite{chakraborty2025biasmap}              & \xmark & \cmark & \cmark & \xmark & \cmark & \xmark \\
Asyrp~\cite{kwon2023asyrp}                         & \xmark & \cmark & \cmark & \xmark & \cmark & \xmark \\
Fair Mapping~\cite{li2025fairmapping}              & \cmark & \xmark & \xmark & \cmark & \xmark & \xmark \\
TIME~\cite{orgad2023time}                          & \cmark & \cmark & \xmark & \xmark & \cmark & \xmark \\
UCE~\cite{gandikota2024uce}                        & \cmark & \cmark & \xmark & \xmark & \cmark & \cmark \\
Finetuning~\cite{shen2024finetuning}               & \cmark & \xmark & \xmark & \cmark & \xmark & \xmark \\
MAS~\cite{zhou2024mas}                             & \cmark & \xmark & \xmark & \cmark & \xmark & \cmark \\
\midrule
\textbf{CO-ALIGN (Ours)}                           & \cmark & \xmark & \xmark & \cmark & \cmark & \cmark \\
\bottomrule
\end{tabular}}
\end{table}

\section{Understanding bias}
As discussed in section \ref{subsec:dataset-bias-and-implication}, the major root cause for biased behavior arises from the inherent bias in the training dataset. Figure \ref{fig:demographic bias}(in appendix) shows the distributional bias across occupations for gender, race and age in the LAION 2B dataset ~\cite{schuhmann2022laionb} (with FairFace classifier ~\citep{9423296}) used for training Stable Diffusion 1.5. 
This bias is then amplified by the model during the training phase due to the tendency of the reverse diffusion process to navigate towards dominant modes ~\cite{seshadri2023bias,perera2023analyzing,perera2023analyzingbiasdiffusionbasedface}, resulting in the generation of biased content.
\begin{figure}
    \centering
    \includegraphics[width=0.7\linewidth]{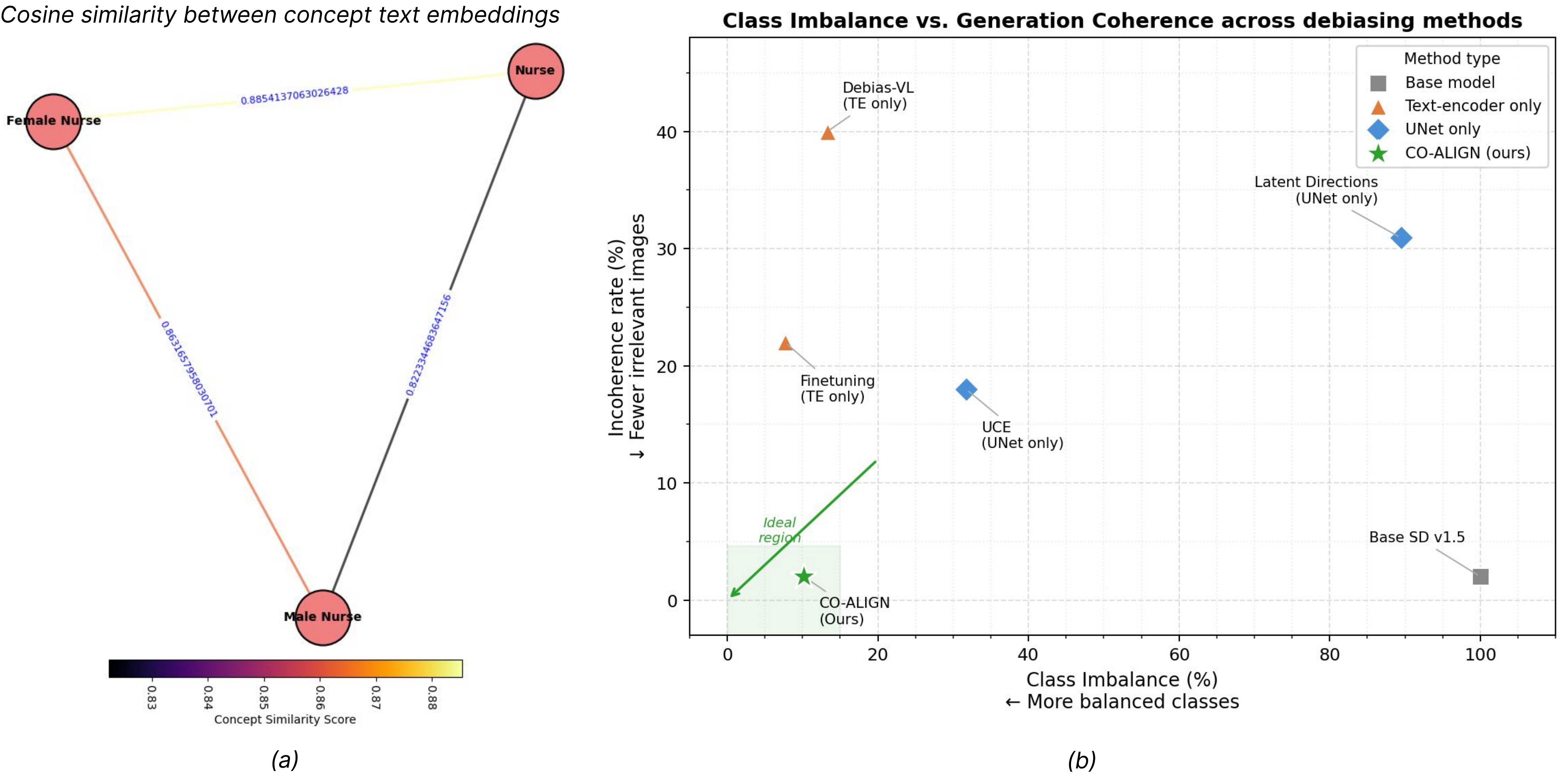}
    \caption{(a) the cosine similarity between the text encoder's embeddings of concepts of Nurse, Female Nurse and Male Nurse; and (b) Figure comparing the class imbalance \% (x-axis: lower is better) against Incoherence rate (y-axis: fraction of generations unclassifiable as the target, lower is better) for five debiasing methods and the base model. 
    }
    \label{fig:cosine-sim-and-scatter}
    \vspace{-10pt}
\end{figure}

\textbf{Is it a Query Understanding or Model Interpretation Problem?}
A fundamental question that divides the bias mitigation literature is whether the observed generative bias is primarily a failure of \emph{query understanding}, \emph{i.e.}, the text encoder misinterprets the input prompt, or a failure of \emph{model interpretation} \emph{i.e.}, the denoiser's internal visual concept representations are themselves skewed, and would produce biased outputs regardless of prompt comprehension.
Several observations from the literature suggest that both components are affected, but in structurally distinct ways.
On the text encoder, underspecified prompts are implicitly enriched with demographic associations~\citep{orgad2023time}, and removing biased directions from text embeddings alone produces measurably fairer outputs~\citep{chuang2023debiasing}. 
Figure \ref{fig:cosine-sim-and-scatter} (a) supports this argument, where we see that 
the cosine similarity between the text encoder's embeddings of \textit{Nurse}, \textit{Female Nurse}, and \textit{Male Nurse} reveals that the model's query understanding of \textit{Nurse} is inherently skewed toward the feminine.

On the denoiser side, specific neuron-level dimensions in the denoiser's h-space are independently responsible for demographic bias~\citep{shi2025difflens}, and gender bias is present in the cross-attention layers regardless of the text conditioning signal~\citep{wu2024revealingbias}. While an analogous diagnostic does not yet exist for the denoiser, we hypothesize that a similar relational concept graph extracted from its cross-attention layers would reveal a structurally similar pattern. We later confirm this hypothesis empirically in Section~\ref{sec:extraction-of-concept-graph}, where \NAME{} extracts exactly such a graph and shows that the denoiser's internal concept topology mirrors the text encoder's bias.


\textbf{Hypothesis: joint alignment is necessary.}
Single-component interventions are suboptimal because the text encoder and denoiser share an entangled geometric representation: correcting one without the other disrupts this alignment, degrading generation coherence even as class imbalance improves.
Figure~\ref{fig:cosine-sim-and-scatter}(b) illustrates this trade-off: text-encoder-only methods (Debias-VL~\citep{chuang2023debiasing}, Finetuning~\citep{shen2024finetuning}) and denoiser-only methods (UCE~\citep{gandikota2024uce}, Latent Directions~\citep{li2024selfdiscovering}) all reduce class imbalance at the cost of a sharp increase in semantically incoherent generations.
We therefore hypothesize that effective debiasing requires \emph{joint} alignment of both components, preserving the geometric harmony between them.
We later confirm this hypothesis empirically: \NAME{} achieves lower class imbalance than all single-component baselines while maintaining generation coherence comparable to the unedited base model.

\section{Extraction of a Concept Graph}
\textbf{Background on Concept Neurons.}
\label{sec:trust}
Prior work~\citep{10.5555/3692070.3692199,fan2024salun,10.5555/3618408.3619298} has established that information about a specific concept is concentrated in a sparse subset of parameters within the CA projection matrices, termed \emph{concept neurons}. Following TRUST~\citep{mansi2026selectivefinetuningtargetedrobust}, we define a parameter $\theta_{c_u}$ as a concept neuron for concept $c_u$ if perturbing its value produces a measurable change in the model's alignment to $c_u$, quantified via CLIP score:
\begin{align}
    \mathcal{L}_{c_u} &= \text{CLIPScore}(I_{c_u},\, c_u), \label{eq:concept-neuron-loss}\\
    \mathcal{M}_r(c_u) &= \eta\!\left(\mathbb{E}\!\left[|\nabla_{\theta=\theta_0}\mathcal{L}_{c_u}|\right] > \gamma\right)^r, \label{eq:concept-neuron-mask}
\end{align}
where $I_{c_u}$ is an image generated conditioned on $c_u$, $\eta(\cdot)$ is an element-wise indicator, and $\gamma = \xi \cdot \sigma_G + \mu_G$ is an adaptive threshold over the gradient matrix $G = |\nabla_{\theta=\theta_0}\mathcal{L}_{c_u}|^r$. \NAME{} uses concept neurons not for unlearning, but to construct a concept-level knowledge graph: nodes represent the cumulative strength of concept neuron populations per concept, and edges represent the shared neuron strength between concepts.

\label{sec:extraction-of-concept-graph}
\begin{figure}
    \centering
    \includegraphics[width=0.98\linewidth]{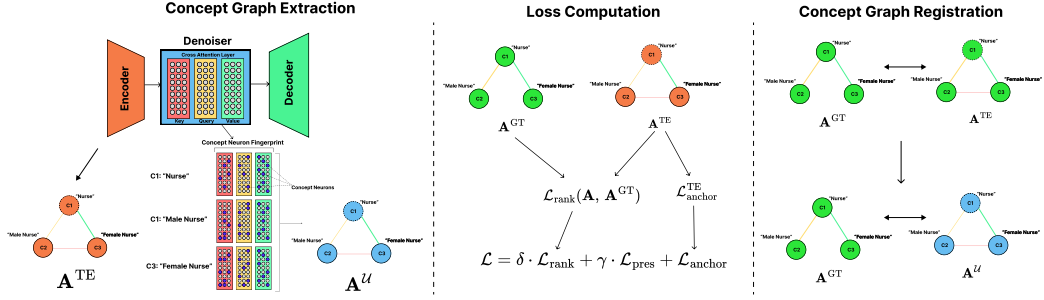}
    \caption{The Figure shows \NAME{}'s three staged finetuning pipeline. In the concept graphs the dotted nodes represent Target Concepts and solid nodes represent the Anchor Concepts.}
    \label{fig:full-pipeline}
    \vspace{-5pt}
\end{figure}

Let $\mathcal{C} = \{c_1, \ldots, c_n\}$ be a concept vocabulary. We represent the concept graph as a weighted, symmetric adjacency matrix $\mathbf{A} \in \mathbb{R}^{n \times n}$, where $A_{ij}$ quantifies the shared neural substrate between $c_i$ and $c_j$. We define such graphs separately for the text encoder (\S\ref{sec:te-kg}) and denoiser (\S\ref{sec:unet-kg}), then jointly align them in \S\ref{sec:bias-mitigation}.

\textbf{Text Encoder.}
\label{sec:te-kg}
The concept representation of $c_i$ is the mean-pooled last hidden state of the text encoder $f_\phi$:
\begin{equation}
    \mathbf{e}_i = \frac{1}{L_i}\sum_{\ell=1}^{L_i} f_\phi(c_i)_\ell \;\in\; \mathbb{R}^d.
    \label{eq:te-embedding}
\end{equation}
The \textit{text-encoder concept graph} $\mathcal{G}^{\mathrm{TE}} = (\mathcal{C},\, \mathbf{A}^{\mathrm{TE}})$ is defined by pairwise cosine similarity of $\ell_2$-normalised embeddings:
\begin{equation}
    A^{\mathrm{TE}}_{ij} = \frac{\mathbf{e}_i^{\top} \mathbf{e}_j}{\|\mathbf{e}_i\|\,\|\mathbf{e}_j\|} = \hat{\mathbf{e}}_i^{\top} \hat{\mathbf{e}}_j, \qquad \mathbf{A}^{\mathrm{TE}} = \hat{E}\,\hat{E}^{\top},
    \label{eq:te-kg}
\end{equation}
where $\hat{E} = [\hat{\mathbf{e}}_1, \ldots, \hat{\mathbf{e}}_n]^{\top} \in \mathbb{R}^{n \times d}$. Off-diagonal entries lie in $[-1, 1]$ and capture directional alignment of concept embeddings.

\textbf{Denoiser.}
\label{sec:unet-kg}
Unlike the text encoder, the denoiser has no single embedding per concept; concept binding is distributed across the cross-attention projection matrices $W_K, W_Q, W_V$. We represent each concept's position in the denoiser's ontology by its \textit{concept neuron fingerprint} --- the masked gradient profile over cross-attention weights.

For concept $c_u$ and projection type $r \in \{k, q, v\}$, the per-head importance of head $h$ in layer $\ell$ is:
\begin{equation}
    G^{\ell,h}_{r}(c_u) = \mathbb{E}\!\left[\,\bigl|\nabla_{W^{\ell,h}_{r}}\mathcal{L}_{c_u}\bigr|\,\right]_{\text{mean}}.
    \label{eq:per-head-importance}
\end{equation}
Stacking over all $L$ layers and $H$ heads yields $\mathbf{G}_r(c_u) \in \mathbb{R}^{LH}$. The masked fingerprint is:
\begin{align}
    \mathcal{M}_r(c_u) &= \eta\!\left(\mathbf{G}_r(c_u) > \gamma\right), \quad \gamma = \xi \cdot \sigma(\mathbf{G}_r(c_u)) + \mu(\mathbf{G}_r(c_u)), \label{eq:mask-flat}\\
    \mathbf{g}_r(c_u) &= \mathbf{G}_r(c_u) \odot \mathcal{M}_r(c_u) \in \mathbb{R}^{LH}. \label{eq:fingerprint}
\end{align}
The \textit{denoiser concept graph} $\mathcal{G}^{\mathcal{U}} = (\mathcal{C},\, \mathbf{A}^{\mathcal{U}})$ is then:
\begin{equation}
    A^{\mathcal{U}}_{ij} = \frac{1}{3} \sum_{r \in \{k,\, q,\, v\}} \cos\!\left(\mathbf{g}_r(c_i),\; \mathbf{g}_r(c_j)\right),
    \label{eq:unet-kg}
\end{equation}
where a high $A^{\mathcal{U}}_{ij}$ indicates that $c_i$ and $c_j$ share disproportionate neural substrate in the denoiser's cross-attention layers.

\section{Concept Graphs for Bias Mitigation}
\label{sec:bias-mitigation}

\textbf{Concept Graph Registration.}
\label{sec:registration}
The target graph $\mathbf{A}^{\mathrm{GT}}$ encodes the desired concept topology. The concept vocabulary is partitioned into a \textbf{target concept} $c^*$ (e.g.\ ``Nurse''), free to adapt during training, and \textbf{anchor concepts} $\mathcal{C}_S$ (e.g.\ \{``Male Nurse'', ``Female Nurse''\}), held fixed via $\mathcal{L}_{\mathrm{anchor}}$. The registration encodes equal proximity of $c^*$ to all attribute concepts:
\begin{equation}
    A^{\mathrm{GT}}_{c^*,\, c} = s^*, \quad \forall\, c \in \mathcal{C}_S,
    \label{eq:bias-registration}
\end{equation}
where $s^* \in [0,1]$ is the target similarity (typically $s^* = 1.0$). The supervised row set is $\mathcal{S} = \{c^*\}$, so $\mathcal{L}_{\mathrm{rank}}$ propagates gradients only through the target concept's row. Alignment is applied sequentially, first to the text encoder, then to the denoiser recalibrated to the updated geometry, as illustrated in Figure~\ref{fig:error repair}.

Now that we have the target graph
$\mathbf{A}^{\mathrm{GT}}$, we define a specialized training objective to achieve its registration with $\mathbf{A}^{\mathrm{TE}}$ and $\mathbf{A}^{\mathcal{U}}$.

\textbf{Training Objective.} Let $\mathbf{A}$ denote either $\mathbf{A}^{\mathrm{TE}}$ or
$\mathbf{A}^{\mathcal{U}}$, and let $\mathbf{A}^{\mathrm{GT}}$ be a target
adjacency matrix encoding the desired concept topology (\S\ref{sec:registration}).
CO-ALIGN aligns $\mathbf{A}$ toward $\mathbf{A}^{\mathrm{GT}}$ via a
differentiable ranking loss that operates row-wise over a supervised subset
$\mathcal{S} \subseteq \mathcal{C}$.
\textbf{Differentiable ranking loss.}
For each concept $c_i \in \mathcal{S}$, the adjacency matrix $\mathbf{A}$ induces a ranking of all concepts by proximity to $c_i$. We encode this via \textit{soft ranks} using differentiable sigmoid pairwise comparisons, and align them to \textit{hard rank targets} $\bar{r}^{\mathrm{GT}}_{ij}$ derived from $\mathbf{A}^{\mathrm{GT}}$:
\begin{align}
    \tilde{r}_{ij} &= \frac{1}{n}\!\left(1 + \sum_{k=1}^{n} \sigma\!\!\left(\frac{A_{ik} - A_{ij}}{\tau}\right)\!\right), \quad \tilde{r}_{ij} \in \left(\tfrac{1}{n},\; 1\right], \label{eq:soft-rank}\\
    \bar{r}^{\mathrm{GT}}_{ij} &= \frac{1}{n}\cdot \operatorname{avgrank}_k\!\!\left(A^{\mathrm{GT}}_{ij}\;\big|\;\text{row}\ i\right), \label{eq:hard-rank}\\
    \mathcal{L}_{\mathrm{rank}} &= \frac{1}{|\mathcal{S}|\cdot n} \sum_{i \in \mathcal{S}}\;\sum_{j=1}^{n} \!\left(\tilde{r}_{ij} - \bar{r}^{\mathrm{GT}}_{ij}\right)^{\!2}, \label{eq:ranking-loss}
\end{align}
where $\tau > 0$ controls sharpness and larger $\tilde{r}_{ij}$ indicates lower similarity rank relative to $c_i$.

\textbf{Static anchor loss.}
A subset $\mathcal{C}_S \subset \mathcal{C}$ of \textit{anchor concepts} retain their pre-alignment representations. For the text encoder and denoiser respectively:
\begin{align}
    \mathcal{L}_{\mathrm{anchor}}^{\mathrm{TE}} &= \sum_{c \in \mathcal{C}_S} \lambda_c\,\bigl\|\mathbf{e}^{(0)}_c - \mathbf{e}_c\bigr\|^2, \label{eq:anchor-te}\\
    \mathcal{L}_{\mathrm{anchor}}^{\mathcal{U}} &= \sum_{c \in \mathcal{C}_S} \lambda_c\,\bigl\|\mathbf{A}^{\mathcal{U},(0)}[c,\,:] - \mathbf{A}^{\mathcal{U}}[c,\,:]\bigr\|^2, \label{eq:anchor-unet}
\end{align}
where $\mathbf{e}^{(0)}_c$ and $\mathbf{A}^{\mathcal{U},(0)}[c,\,:]$ are the embedding and CG row of $c$ under the frozen base model.

\textbf{Preservation loss.}
To prevent catastrophic forgetting of concepts outside $\mathcal{C}$, we augment with the standard LDM noise-prediction loss on reference pairs $\mathcal{D}_{\mathrm{pres}}$:
\begin{equation}
    \mathcal{L}_{\mathrm{pres}} = \mathbb{E}_{(x_0,\tilde{c}),\epsilon,t}\! \left[ \left\| \epsilon - \epsilon_\theta\!\left( \sqrt{\bar{\alpha}_t}\,z_0 + \sqrt{1-\bar{\alpha}_t}\,\epsilon,\; f_\phi(\tilde{c}),\; t \right) \right\|^2 \right], \label{eq:pres-loss}
\end{equation}
where $z_0 = \mathcal{E}(x_0)$ is the VAE-encoded latent.

\textbf{Total objective.}
\begin{equation}
    \mathcal{L} = \delta\cdot\mathcal{L}_{\mathrm{rank}} + \gamma\cdot\mathcal{L}_{\mathrm{pres}} + \mathcal{L}_{\mathrm{anchor}}, \label{eq:total-loss}
\end{equation}
where $\delta, \gamma \geq 0$ balance the relative contributions.

\section{Experiments and Results}
\label{sec:experiments}

We now empirically evaluate our method, assessing both its effectiveness at mitigating bias and its impact on generation quality.

\textbf{Experimental Setup}
\label{sec:setup}

\textbf{Models and datasets.}
All experiments use Stable Diffusion v1.5~\citep{9878449} as the base
generative model.  For the debiasing evaluation we follow the protocol of
DiffLens~\citep{shi2025difflens}: we generate $500$ images per occupation prompt
(\textit{``face of a/an \{occupation\}''}) across four occupations, namely,
\textit{nurse}, \textit{doctor}, \textit{fire fighter}, and \textit{receptionist}
, and evaluate three demographic axes: gender (Male/Female), race
(White, Black, Asian, Indian), and age (Young,Adult,Old).
Demographic labels are obtained from the FairFace~\citep{9423296} 4-race
ResNet-34 classifier.  

\textbf{Baselines and metrics.}
We compare against five state-of-the-art debiasing baselines: \textbf{Asyrp}~\citep{kwon2023diffusion},
\textbf{H-Distribution}~\citep{parihar2024balancing}, \textbf{Latent
Directions}~\citep{li2024selfdiscovering}, \textbf{Finetuning}~\citep{shen2024finetuning}, and
\textbf{DiffLens}~\citep{shi2025difflens}.  
We compute four metrics for comparing the fairness, and generation quality of \NAME{} against the baselines. We compute Fairness Discrepancy (FD)~\citep{parihar2024balancing} for evaluating the fairness of debiasing, FID~\citep{heusel2017fid} for evauating the photorealism, and CLIP-I, and CLIP-T scores to evaluate the fidelity of generation. We discuss these metrics in detail in Appendix \ref{appndx:evalmetrics}

\textbf{Evaluation Results}
\label{sec:results}

\textbf{Fairness.}
\NAME{} achieves the lowest FD for gender ($\text{FD}=0.032$, $30.4\%$ improvement over the next best) and race ($\text{FD}=0.043$, $75.4\%$ improvement over the next best), and a competitive FD for age ($\text{FD}=0.054$), on par with DiffLens.

\textbf{Generation quality.}
On image quality \NAME{} achieves $\Delta\mathrm{FID} = 11.4$ below the base
model average, outperforming all baselines on FID.  CLIP-I scores of $0.9506$ (gender),
$0.9525$ (age), $0.9122$ (race) confirm that the aligned model's outputs
remain close to those of the base model, representing a $2.8\%$ improvement in
image fidelity over the strongest baseline.  CLIP-T scores are stable across
all methods, indicating that prompt adherence is preserved throughout alignment, and is the best across all baselines.

\begin{figure}
    \centering
    \includegraphics[width=0.75\linewidth]{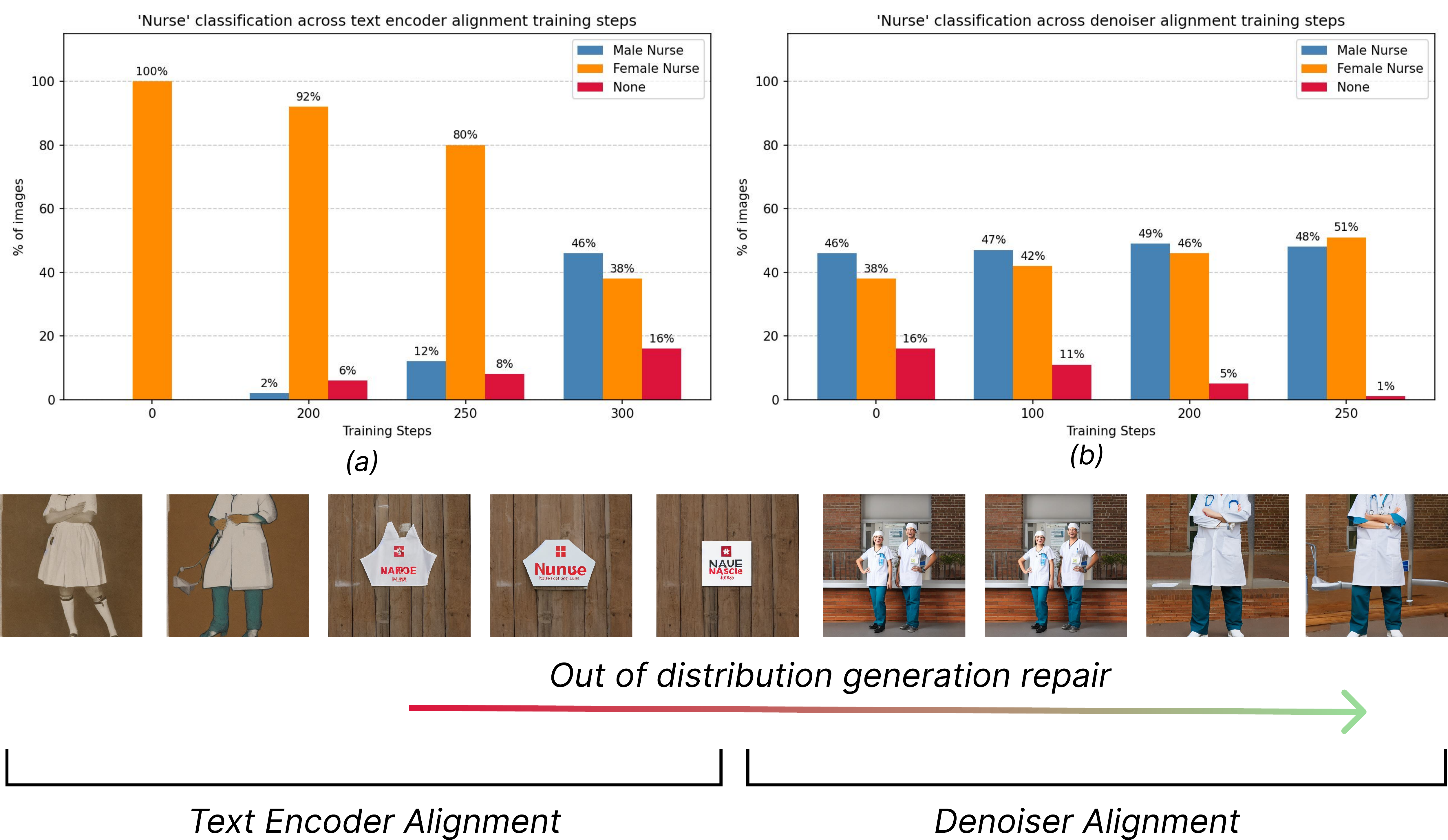}
    \caption{The figure shows how alignment of just the text encoder leads to generation of incoherent output due to mismatch in representations between the text encoder and the denoising unit. The subsequent alignment of the denoiser reconfigures the representation geometry between the text encoder and the denoiser, thus regaining coherence.}
    \label{fig:error repair}
    \vspace{-10pt}
\end{figure}

\textbf{Incoherence.} A distinctive failure mode of single-component debiasing is \textit{semantic incoherence}: generations that fail to depict the target concept, arising from the distribution shift between the edited and unedited components, measured as the fraction of generated images classified by LLaVA as off-target.
Figure~\ref{fig:cosine-sim-and-scatter}(b) shows that both text-encoder-only (Debias-VL, Finetuning) and denoiser-only (UCE, Latent Directions) methods trade fairness for incoherence, while \NAME{} is the only method in the ideal bottom-left region, reducing incoherence by \textbf{88\%} over the base model while achieving the best fairness.
Figure~\ref{fig:error repair} further illustrates this: text-encoder alignment alone produces incoherent outputs, which denoiser alignment resolves by reconfiguring the denoiser's concept graph to match the updated embedding geometry.

\begin{table}[t]
\centering
\caption{Evaluation of bias mitigation in text-to-image diffusion model Stable Diffusion~\cite{9878449}, based on average performance across four occupations. We highlight the best results in \textbf{bold} and the second-best with \underline{underlined} text (excluding the ``original'').}
\label{tab:bias_mitigation-baseline-comp}
\resizebox{\textwidth}{!}{%
\begin{tabular}{l cccc cccc cccc}
\toprule
\multirow{2}{*}{\textbf{Method}}
  & \multicolumn{4}{c}{\textbf{Gender (2)}}
  & \multicolumn{4}{c}{\textbf{Age (3)}}
  & \multicolumn{4}{c}{\textbf{Race (4)}} \\
\cmidrule(lr){2-5} \cmidrule(lr){6-9} \cmidrule(lr){10-13}
  & \textbf{FD} $\downarrow$ & \textbf{FID} $\downarrow$ & \textbf{CLIP-I} $\uparrow$ & \textbf{CLIP-T} $\uparrow$
  & \textbf{FD} $\downarrow$ & \textbf{FID} $\downarrow$ & \textbf{CLIP-I} $\uparrow$ & \textbf{CLIP-T} $\uparrow$
  & \textbf{FD} $\downarrow$ & \textbf{FID} $\downarrow$ & \textbf{CLIP-I} $\uparrow$ & \textbf{CLIP-T} $\uparrow$ \\
\midrule
Original
  & 0.564 & 120.06 & --     & 0.6155
  & 0.752 & 120.06 & --     & 0.6155
  & 0.558 & 120.06 & --     & 0.6155 \\
Asyrp~\cite{kwon2023diffusion}
  & 0.408 & 166.11 & 0.8253 & 0.6005
  & 0.682 & 200.90 & 0.8527 & \textbf{0.6122}
  & 0.524 & 153.06 & \underline{0.8804} & 0.6086 \\
H-Distribution~\cite{parihar2024balancing}
  & 0.222 & 151.68 & 0.8475 & 0.6087
  & 0.506 & 147.71 & 0.8345 & \underline{0.6098}
  & 0.544 & 126.90 & 0.8255 & 0.6100 \\
Latent Direction~\cite{li2024selfdiscovering}
  & 0.305 & \underline{129.37} & 0.8058 & \underline{0.6091}
  & \underline{0.052} & \underline{113.81} & 0.8151 & 0.6067
  & \textbf{0.175} & 128.30 & 0.8211 & \underline{0.6132} \\
Finetuning~\cite{shen2024finetuning}
  & \underline{0.050} & 161.47 & \underline{0.8779} & \underline{0.6095}
  & 0.746 & 161.47 & \underline{0.8779} & 0.6095
  & \underline{0.198} & 161.47 & 0.8779 & 0.6095 \\
DiffLens~\cite{shi2025difflens}
  & \underline{0.046} & \underline{112.83} & 0.8501 & 0.6090
  & \textbf{0.049} & \underline{99.17}  & 0.8778 & 0.6057
  & 0.401 & \underline{119.86} & \underline{0.9096} & \underline{0.6149} \\

UCE~\cite{gandikota2024uce}
  & 0.4117 & 132.60 & 0.7874 & 0.6155
  & 0.5792 & 122.86  & 0.9169 & 0.6136
  & 0.2073 & 110.94 & 0.9338 & 0.6144 \\

\midrule
\textbf{\NAME{} (Ours)}
  & \textbf{0.032} & \textbf{98.88} & \textbf{0.9506} & \textbf{0.6153}
  & 0.054 & \textbf{97.42} & \textbf{0.9525} & \textbf{0.6192}
  & \textbf{0.043} & \textbf{101.45} & \textbf{0.9122} & \textbf{0.6219} \\
\bottomrule
\end{tabular}%
}
\end{table}

\section{Beyond Bias Mitigation: a Case Study on Unlearning}
\label{sec:beyond}



Post-hoc concept unlearning techniques erase a target concept $c_T$ by editing model parameters~\citep{gandikota2023erasing,gandikota2024uce} or steering its latent representation~\citep{yoon2025safree}. However, only concepts within a \textit{locality radius} of $c_T$ in concept space are erased; adversarial bypass concepts outside this neighbourhood $\mathcal{N}(c_T)$ remain unaffected, allowing attackers to recover unlearned content via semantically adjacent prompts.

\begin{wraptable}{r}{0.7\textwidth}
\vspace{-14pt}
\centering
\caption{Percentage of images flagged as nudity by VLM classifier under adversarial prompt categories (\textbf{lower is better}). \textbf{Bold} denotes best result per column within each group. \NAME{} pre-aligns adversarial concepts closer to nudity in the model's concept space, substantially boosting the effectiveness of post-hoc unlearning.}
\label{tab:adversarial_robustness}
\setlength{\tabcolsep}{3pt}
\renewcommand{\arraystretch}{0.95}
\scriptsize
\begin{tabular}{llcccccr}
\toprule
\textbf{Base Model} & \makecell{\textbf{Unlearning}\\\textbf{Technique}} & \makecell{\textbf{Porn}\\\textbf{Star}} & \textbf{Nymph.} & \textbf{Cream.} & \textbf{Shirt.} & \makecell{\textbf{Att.}\\\textbf{Fem.}} & \textbf{Avg} \\
\midrule
\multirow{4}{*}{SD v1.5}
  & None        & 14.0 & 62.0  & 100.0          & 40.0         & 26.0         & 48.4 \\
  & $+$ UCE     &  2.0 & 40.0  &  88.0          & 10.0         &  8.0         & 29.6 \\
  & $+$ ESD     &  2.0 & 14.0  &  \textbf{12.0} & 18.0         &  4.0         & \textbf{10.0} \\
  & $+$ SAFREE  &  \textbf{0.0} & \textbf{0.0} & 94.0 & \textbf{0.0} & \textbf{0.0} & 18.8 \\
\midrule
\multirow{4}{*}{\makecell[l]{SD v1.5\\$+$\textbf{\NAME{}}\\(Ours)}}
  & None        & 16.0 & 100.0         & 100.0         & 60.0         & 26.0         & 60.4 \\
\cdashline{2-8}[2pt/2pt]
  & $+$ UCE     & \textbf{0.0} & 18.0  &  32.0         & \textbf{0.0} & \textbf{0.0} & 10.0 \\
  & $+$ ESD     & \textbf{0.0} & \textbf{0.0} & 2.0   &  2.0         & \textbf{0.0} & \textbf{0.8} \\
  & $+$ SAFREE  & \textbf{0.0} &  4.0  & \textbf{0.0}  &  2.0         & \textbf{0.0} & 1.2 \\
\bottomrule
\end{tabular}
\vspace{-6pt}
\end{wraptable}
We exploit \NAME{}'s concept topology editing to address this vulnerability. Before unlearning, we use \NAME{} to pull a set of known adversarial concepts $\mathcal{C}_{\mathrm{adv}}$ (e.g., for $c_T = \textit{Nudity}$: $\mathcal{C}_{\mathrm{adv}} = \{\textit{Nymphettes},\, \textit{Creampie}\}$) closer to $c_T$ in both the text encoder and denoiser concept graphs, bringing them within $\mathcal{N}(c_T)$ before any unlearning technique is applied.

\textbf{Neighbourhood Pulling Effect}
\label{sec:neighbourhood}

A key emergent property of concept graph alignment is \textit{semantic
neighbourhood propagation}: aligning a supervised concept $c^* \in
\mathcal{C}_{\mathrm{adv}}$ toward $c_T$ causes its unsupervised semantic
neighbors to shift toward $c_T$ as well, without any direct supervision
signal.  Formally, let $\mathcal{N}_\epsilon(c^*)$ denote the set of
concepts within concept distance $\epsilon$ of $c^*$ in the model's concept
graph before alignment.  After aligning $c^*$ toward $c_T$ via
$\mathcal{L}_{\mathrm{rank}}$, we observe that for all
$c_n \in \mathcal{N}_\epsilon(c^*)$:
\begin{equation}
    \Delta |c_n,\, c_T|_{CG}
    = |c_n, c_T|_{CG_{{\mathrm{after}}}} - |c_n, c_T|_{CG_{{\mathrm{before}}}} > 0,
    \label{eq:neighbourhood-pull}
\end{equation}
even though $c_n$ received no direct alignment supervision.  This effect is
analogous to label propagation in graph-based semi-supervised
learning~\citep{zhou2003learning,zhu2003semi,iscen2019label}: the graph smoothness assumption
propagates alignment from supervised nodes to adjacent unsupervised ones.

\begin{figure}
    \centering
    \includegraphics[width=0.9\linewidth]{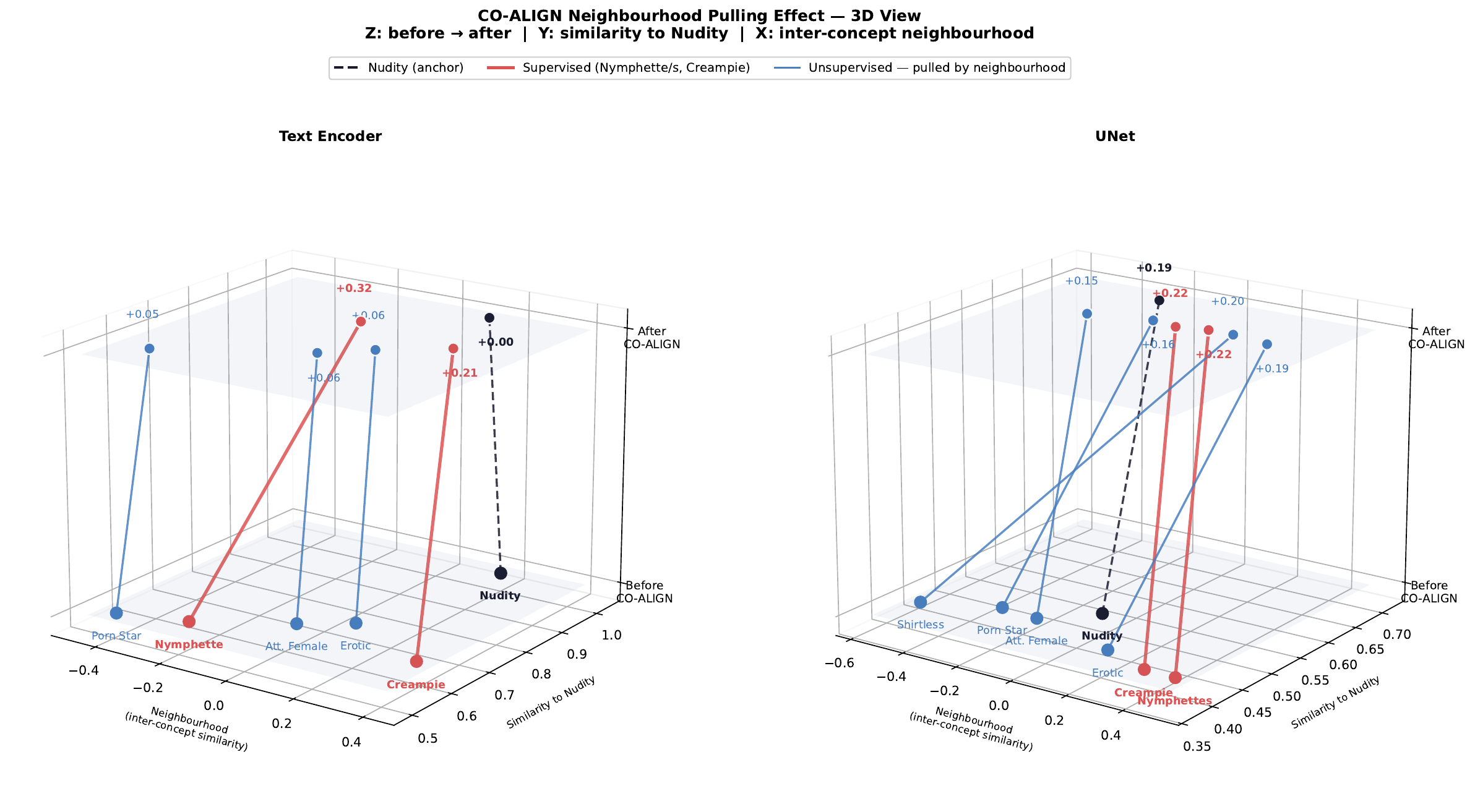}
    \caption{The figure shows the Semantic Neighborhood Pulling Effect on Concept Graphs as a result of \NAME{} edit for both the Text encoder and the Unet. The Y-axis measures distance to Nudity; the X-axis reflects inter-concept neighbourhood structure via 1-D MDS (concepts closer in embedding space appear closer on X).}
    \label{fig:neughbourhood-pull}
    \vspace{-8pt}
\end{figure}

Figure~\ref{fig:neughbourhood-pull} visualizes this effect for both the text
encoder and the UNet.  Each line traces a concept's similarity to Nudity
before (z$=0$) and after (z$=1$) \NAME{} alignment.  Red lines denote the
two directly supervised concepts (Nymphettes, Creampie); blue lines denote
unsupervised concepts (Erotic, Attractive Female, Jonny Sins, Shirtless).
Despite receiving no direct supervision, all unsupervised concepts exhibit a
measurable upward shift in similarity to Nudity ($+0.15$ to $+0.22$ in the
UNet graph), confirming that alignment of $\mathcal{C}_{\mathrm{adv}}$
broadcasts into the surrounding concept neighborhood.

\textbf{\NAME{} for Robust Unlearning}
\label{sec:robust-unlearning}

For the unlearning evaluation, we generate $500$ images per adversarial concept and classify nudity presence with LLaVA~\citep{liu2023visualinstructiontuning}, following the prominent adversarial concepts from I2P dataset ~\citep{10205305}. We evaluate
three unlearning methods \textbf{UCE}~\citep{gandikota2024uce},
\textbf{ESD}~\citep{gandikota2023erasing}, and \textbf{SAFREE}~\citep{yoon2025safree}
applied both to the base model and to a \NAME{}-aligned model.

\textbf{Procedure.}
The CO-ALIGN pre-alignment brings $\mathcal{C}_{\mathrm{adv}}$ and, via semantic neighborhood propagation, a broader set of unsupervised bypass concepts
within the locality radius $\mathcal{N}(c_T)$ before unlearning is applied.
Any subsequent unlearning technique that targets $c_T$ therefore also
covers the pre-aligned bypass concepts, and their neighboring concepts, without requiring knowledge of those
concepts at unlearning time.

\textbf{Robustness of unlearning with \NAME{}.}
Table~\ref{tab:adversarial_robustness} reports the percentage of images flagged as nudity by the VLM classifier under five adversarial prompt categories.  
Across all three unlearning techniques, pre-aligning with \NAME{} substantially improves adversarial robustness.  \NAME{} + ESD reduces the average nudity rate from
$10.0\%$ to $\mathbf{0.8\%}$; \NAME{} + SAFREE from $18.8\%$ to
$\mathbf{1.2\%}$; and \NAME{} + UCE from $29.6\%$ to $\mathbf{10.0\%}$.
Notably, the Creampie prompt which achieves $88\%$-$100\%$ nudity
generation against all baselines, drops to $0\%$-$32\%$ after \NAME{}
pre-alignment, demonstrating that neighborhood propagation successfully
relocates this adversarial bypass concept into the unlearning target's erasure zone.

\section{Conclusion}

We presented \NAME{}, a bias mitigation framework that jointly aligns concept graphs extracted from the text encoder and denoiser toward a target topology. Our central hypothesis, that effective debiasing requires geometric harmony between both components, is confirmed empirically: single-component methods lie off the Pareto frontier of fairness and coherence, while \NAME{} achieves a $30\%$ fairness improvement, $\Delta\text{FID}=11.4$, $2.8\%$ higher image fidelity, and an $88\%$ reduction in semantically incoherent outputs, all without added inference cost. Beyond debiasing, the emergent neighborhood propagation effect improves adversarial robustness of post-hoc unlearning by up to $12.5\times$ without modifying the unlearning procedure. Limitations include the compute cost of denoiser graph extraction and the need to re-derive concept graph construction for transformer-based architectures such as DiT~\citep{10377858} and SDXL~\citep{podell2024sdxl}, which we leave for future work.

\bibliographystyle{plainnat}
\bibliography{references}
\appendix
\section{Inherent Biases in T2I Diffusion Models}
\begin{wrapfigure}[19]{r}{0.45\textwidth}
    \vspace{-30pt}
    \centering
    \includegraphics[width=\linewidth]{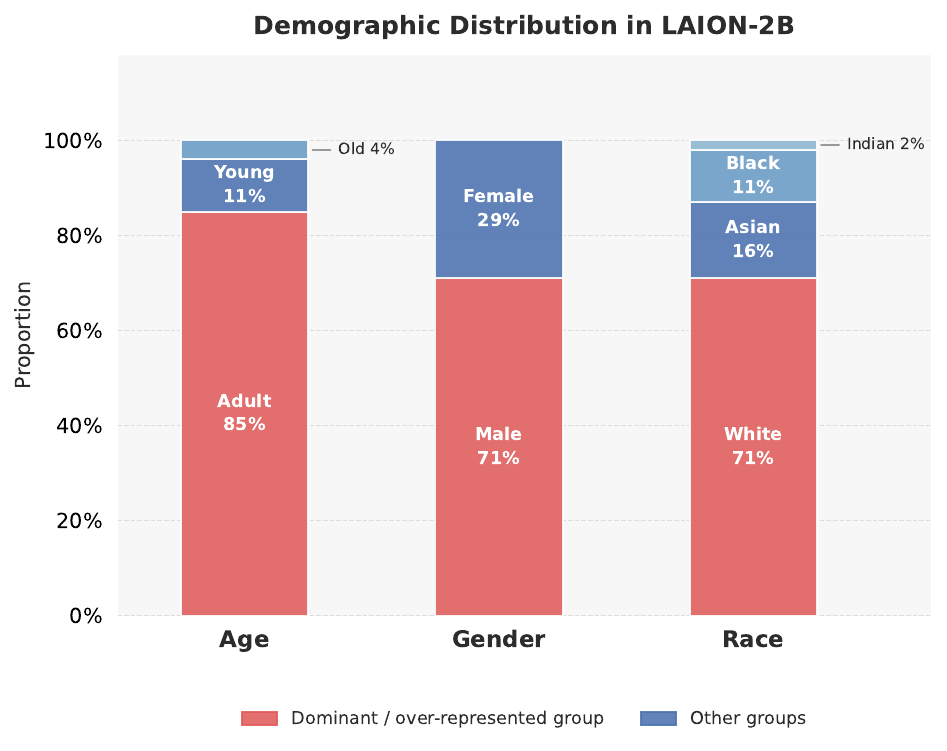}
    \caption{Demographic bias on randomly sampled 10k images from LAION 2B dataset.}
    \label{fig:demographic bias}
\end{wrapfigure}
\textbf{Dataset biases and their implications.}\label{subsec:dataset-bias-and-implication} Alike other generative models, biases in T2I diffusion models often originate from the large-scale web-scraped corpora used for training. LAION-5B~\citep{schuhmann2022laion}, used for training Stable Diffusion, mirrors societal inequalities at internet scale across profession~\citep{bianchi2023easily}, race~\citep{cho2023dall,luccioni2023stable} and culture~\citep{naik2023social}. These biases have shown to be \emph{amplified} beyond the degree present in training data both during model training~\citep{seshadri2023bias,perera2023analyzing} and inference ~\citep{roos2026bias}. Large scale deployment of these models in both human operated and automated systems have shown detrimental societal consequences of reinforcement of stereotypes~\citep{bianchi2023easily}, across gender, age, race, and geography simultaneously~\citep{naik2023social}.

\textbf{Bias mitigation techniques.} Existing debiasing techniques can be broadly classified in two groups, depending on where in the generation pipeline they intervene, as summarised in Table~\ref{tab:related_works}. 

\textit{Text encoder debiasing} methods intervene solely at the text encoder.
Shen et al.~\citep{shen2024finetuning} fine-tune the text encoder directly using a distributional alignment loss on generated images.
Fair Mapping~\citep{li2025fairmapping} learns a lightweight linear remap over conditioning embeddings to project them into a debiased subspace.
Chuang et al.~\citep{chuang2023debiasing} achieve a similar effect training-free, via a calibrated projection matrix that removes biased directions from text embeddings at inference time.
Fair Diffusion~\citep{friedrich2023fairdiffusion} via SEGA~\citep{brack2023sega}, FairGen~\citep{kang2025fairgen}, and Kim et al.~\citep{kim2025rethinking} take a softer approach, steering the conditioning signal or noise initialisation at inference without any weight modification.
While efficient, all text-encoder-only methods leave the denoiser's internal concept organization and the visual biases encoded. 

\textit{Denoiser debiasing} methods instead target just the denoiser's internal representations. TIME~\citep{orgad2023time} edits the key and value projection matrices in the UNet's cross-attention layers in closed form, rerouting underspecified prompts toward target attribute embeddings.
UCE~\citep{gandikota2024uce} extends this to performing multiple edits while only targeting the value projection matrices.
Asyrp~\citep{kwon2023asyrp} establishes the UNet bottleneck (h-space) as a semantically linear latent space amenable to controllable editing.
Building on this, Balancing Act~\citep{parihar2024balancing} train a lightweight Attribute Distribution Predictor on h-space features to guide debiased generation without weight modification, while Li et al.~\citep{li2024selfdiscovering} discover interpretable fairness-sensitive directions in h-space without external classifiers.
Vardhana et al.~\citep{vardhana2026selfdebias} perform fully unsupervised debiasing by clustering image encoder embeddings and steering denoising toward a uniform attribute distribution.
SCALEX~\citep{zeng2026scalex} maps the conceptual structure of diffusion models in h-space via prompt-aligned latent directions derived from Latent Consistency Models.
DIFFLENS~\citep{shi2025difflens} takes a more mechanistic approach, decomposing h-space activations via sparse autoencoders to identify and suppress specific neuron-level dimensions responsible for bias.
BiasMap~\citep{chakraborty2025biasmap} and EFA~\citep{park2025efa} further reveal that many denoiser-side corrections reduce output distributional gaps without disentangling the underlying concept coupling in cross-attention, leaving residual structural bias intact.
Yasser et al.~\citep{yasser2026clip} extend this mechanistic analysis to the CLIP vision encoder, identifying specific attention heads whose ablation reduces gender bias more surgically than age bias.
MAS~\citep{zhou2024mas} addresses the related problem of association-engendered stereotypes arising from the co-generation of multiple concepts.
While these methods edit the denoiser's internal representations, they are applied independently of the text encoder, leaving the biased embedding geometry of underspecified prompts uncorrected.

These two lines of work are thus complementary but incomplete: text-encoder methods leave visual biases in the denoiser intact, while denoiser-side methods inherit a biased conditioning signal. This motivates a global approach that jointly debiases both components.


\section{Algorithms}\label{appendix:algorithm}

\begin{algorithm}[t]
\caption{Text Encoder Concept Graph Extraction}
\label{alg:te-kg}
\begin{algorithmic}[1]
\Require Concept set $\mathcal{C} = \{c_1, \ldots, c_n\}$,
         text encoder $f_\phi$
\Ensure  Adjacency matrix $\mathbf{A}^{\mathrm{TE}} \in \mathbb{R}^{n \times n}$
\For{$i = 1$ \textbf{to} $n$}
    \State Tokenise $c_i$ and compute hidden states via $f_\phi$
    \State $\mathbf{e}_i \leftarrow \text{MeanPool}(\text{LastHiddenState}(f_\phi(c_i)))$
    \Comment{Eq.~\ref{eq:te-embedding}}
    \State $\hat{\mathbf{e}}_i \leftarrow \mathbf{e}_i \,/\, \|\mathbf{e}_i\|_2$
\EndFor
\State Assemble $\hat{E} \leftarrow [\hat{\mathbf{e}}_1, \ldots, \hat{\mathbf{e}}_n]^{\top}$
\State $\mathbf{A}^{\mathrm{TE}} \leftarrow \hat{E}\,\hat{E}^{\top}$
\Comment{$A^{\mathrm{TE}}_{ij} = \hat{\mathbf{e}}_i^{\top}\hat{\mathbf{e}}_j$, Eq.~\ref{eq:te-kg}}
\State \Return $\mathbf{A}^{\mathrm{TE}}$
\end{algorithmic}
\end{algorithm}

\begin{algorithm}[t]
\caption{Denoiser Concept Graph Extraction}
\label{alg:unet-kg}
\begin{algorithmic}[1]
\Require Concept set $\mathcal{C} = \{c_1, \ldots, c_n\}$,
         denoiser $\epsilon_\theta$ with $L$ cross-attention layers
         ($H$ heads each), sensitivity $\xi$
\Ensure  Adjacency matrix $\mathbf{A}^{\mathcal{U}} \in \mathbb{R}^{n \times n}$
\For{$u = 1$ \textbf{to} $n$}
    \State Generate image $I_{c_u}$ by running the full denoising chain conditioned on $c_u$
    \State Compute CLIP alignment loss $\mathcal{L}_{c_u} = \mathrm{CLIPScore}(I_{c_u}, c_u)$
    \Comment{Eq.~\ref{eq:concept-neuron-loss}}
    \For{$r \in \{k, q, v\}$}
        \For{$\ell = 1$ \textbf{to} $L$, $h = 1$ \textbf{to} $H$}
            \State $G^{\ell,h}_{r}(c_u) \leftarrow \mathrm{mean}\!\left(|\nabla_{W^{\ell,h}_{r}} \mathcal{L}_{c_u}|\right)$
            \Comment{Eq.~\ref{eq:per-head-importance}}
        \EndFor
        \State Flatten: $\mathbf{G}_r(c_u) \leftarrow [G^{1,1}_r(c_u), \ldots, G^{L,H}_r(c_u)]^{\top}$
        \State $\gamma_r \leftarrow \xi\cdot\sigma(\mathbf{G}_r(c_u)) + \mu(\mathbf{G}_r(c_u))$
        \Comment{Adaptive threshold, Eq.~\ref{eq:mask-flat}}
        \State $\mathcal{M}_r(c_u) \leftarrow \mathbf{1}[\mathbf{G}_r(c_u) > \gamma_r]$
        \State $\mathbf{g}_r(c_u) \leftarrow \mathbf{G}_r(c_u) \odot \mathcal{M}_r(c_u)$
        \Comment{Fingerprint, Eq.~\ref{eq:fingerprint}}
    \EndFor
\EndFor
\For{$i = 1$ \textbf{to} $n$, $j = 1$ \textbf{to} $n$}
    \State $A^{\mathcal{U}}_{ij} \leftarrow \dfrac{1}{3}\displaystyle\sum_{r \in \{k,q,v\}}
           \cos\!\left(\mathbf{g}_r(c_i),\; \mathbf{g}_r(c_j)\right)$
           \Comment{Eq.~\ref{eq:unet-kg}}
\EndFor
\State \Return $\mathbf{A}^{\mathcal{U}}$
\end{algorithmic}
\end{algorithm}

\begin{algorithm}[t]
\caption{Concept Graph Alignment (Text Encoder or Denoiser)}
\label{alg:alignment}
\begin{algorithmic}[1]
\Require Concept set $\mathcal{C}$, target graph $\mathbf{A}^{\mathrm{GT}}$,
         supervised rows $\mathcal{S}$, anchor concepts $\mathcal{C}_S$,
         anchor weights $\{\lambda_c\}_{c \in \mathcal{C}_S}$,
         loss weights $\delta, \gamma$, temperature $\tau$,
         training steps $T$
\Ensure  Aligned model parameters $\theta^*$
\State Freeze parameters of the component \emph{not} being aligned
\State Capture original anchor representations:
    \For{$c \in \mathcal{C}_S$} $\mathbf{a}^{(0)}_c \leftarrow$ current embedding or CG-row of $c$ \EndFor
\For{$t = 1$ \textbf{to} $T$}
    \State Extract current CG $\mathbf{A}$ (Alg.~\ref{alg:te-kg} or~\ref{alg:unet-kg}) retaining the computation graph
    \For{$i \in \mathcal{S}$, $j = 1$ \textbf{to} $n$}
        \State Compute soft rank $\tilde{r}_{ij} \leftarrow
               \tfrac{1}{n}\!\left(1 + \sum_{k} \sigma\!\left(\tfrac{A_{ik} - A_{ij}}{\tau}\right)\!\right)$
               \Comment{Eq.~\ref{eq:soft-rank}}
        \State Compute hard rank target $\bar{r}^{\mathrm{GT}}_{ij}$ from $\mathbf{A}^{\mathrm{GT}}$
               with average-rank tie-breaking
               \Comment{Eq.~\ref{eq:hard-rank}}
    \EndFor
    \State $\mathcal{L}_{\mathrm{rank}} \leftarrow
           \dfrac{1}{|\mathcal{S}|\cdot n}\displaystyle\sum_{i\in\mathcal{S}}\sum_{j}
           \!\left(\tilde{r}_{ij} - \bar{r}^{\mathrm{GT}}_{ij}\right)^{\!2}$
           \Comment{Eq.~\ref{eq:ranking-loss}}
    \State $\mathcal{L}_{\mathrm{anchor}} \leftarrow
           \displaystyle\sum_{c \in \mathcal{C}_S}
           \lambda_c\,\|\mathbf{a}^{(0)}_c - \mathbf{a}_c\|^2$
           \Comment{Eq.~\ref{eq:anchor-te} or~\ref{eq:anchor-unet}}
    \State $\mathcal{L}_{\mathrm{pres}} \leftarrow
           \mathbb{E}_{(x_0,\tilde{c})\sim\mathcal{D}_{\mathrm{pres}},\,\epsilon,\,t}
           \!\left[\|\epsilon - \epsilon_\theta(z_t, f_\phi(\tilde{c}), t)\|^2\right]$
           \Comment{Eq.~\ref{eq:pres-loss}}
    \State $\mathcal{L} \leftarrow
           \delta\cdot\mathcal{L}_{\mathrm{rank}}
           + \gamma\cdot\mathcal{L}_{\mathrm{pres}}
           + \mathcal{L}_{\mathrm{anchor}}$
           \Comment{Eq.~\ref{eq:total-loss}}
    \State $\theta \leftarrow \theta - \eta\,\nabla_\theta\,\mathcal{L}$
\EndFor
\State \Return $\theta$
\end{algorithmic}
\end{algorithm}

\section{Background on T2I Diffusion Models}
\label{sec:background}

\textbf{T2I Diffusion Models}
Latent diffusion models~\citep{9878449} operate by learning to invert a fixed forward noising process.
Given a data sample $x_0$, a forward process corrupts it into Gaussian noise $x_T$ over $T$ timesteps via:
\begin{equation}
    q(x_t | x_{t-1}) := \mathcal{N}\!\left(x_t;\, \sqrt{\alpha_t}\, x_{t-1},\, (1-\alpha_t)\mathbf{I}\right),
    \label{eq:t2i-diffusion}
\end{equation}
where $\alpha_t \in (0,1)$ is the variance schedule. A denoiser $\epsilon_\theta$, parameterised as a conditional UNet, is trained to predict the added noise:
\begin{equation}
    \mathcal{L}_{\text{LDM}} = \mathbb{E}_{z_t,\epsilon,t,c}\left[\|\epsilon - \epsilon_\theta(z_t, c, t)\|^2\right],
    \label{eq:ldm}
\end{equation}
where $z_t$ is the noisy latent at timestep $t$ and $c$ is a text embedding obtained from a pretrained CLIP text encoder~\citep{DBLP:conf/icml/RadfordKHRGASAM21}.

\textbf{Cross-Attention.}
Semantic alignment between text and image is mediated by cross-attention (CA) at each layer of the denoiser.
Given image-derived queries $Q \in \mathbb{R}^{N \times d}$ and text-derived keys $K \in \mathbb{R}^{L \times d}$, values $V \in \mathbb{R}^{L \times d}$:
\begin{equation}
    \text{Cross Attention}(Q, K, V) = \text{softmax}\!\left(\frac{QK^\top}{\sqrt{d}}\right)V,
    \label{eq:ca}
\end{equation}
where $N$ is the number of spatial locations in $z_t$ and $L$ is the number of text tokens.
$K$ and $V$ are linear projections of the text embedding $c$, making the CA layers the primary site of text-visual concept binding~\citep{orgad2023time,gandikota2024uce}.
The projection matrices $W_K$ and $W_V$ therefore encode the model's concept representations and are the principal target of both bias mitigation and concept unlearning methods.

\section{Evaluation Metrics}
\label{appndx:evalmetrics}
We use the following metrics for the bias mitigation accessment:
\begin{itemize}
    \item \textbf{Fairness Discrepancy (FD)}~\cite{parihar2024balancing}: deviation of the
    predicted demographic distribution from the uniform target,
    $\mathrm{FD} = \sqrt{\sum_i (\bar{p}_i - 1/K)^2}$, where $\bar{p}_i$ is
    the mean predicted probability for class $i$ and $K$ is the number of
    classes.  Lower is fairer.
    \item \textbf{FID}~\cite{heusel2017fid}: Fréchet Inception Distance against
    FFHQ as the reference distribution.  Lower is better.
    \item \textbf{CLIP-I} (image fidelity): mean cosine similarity between
    CLIP~\cite{DBLP:conf/icml/RadfordKHRGASAM21} image embeddings of generated and base-model
    reference images, normalised to $[0,1]$.  Higher indicates better
    content preservation.
    \item \textbf{CLIP-T} (text alignment): mean cosine similarity between
    CLIP image and text embeddings for the generation prompt, normalised to
    $[0,1]$.  Higher indicates better semantic fidelity to the prompt.
\end{itemize}

\section{Reproducibility Statement: Hyperparameters, Implementation Details and Compute}
\paragraph{Implementation details.}
Both stages of \NAME{} use AdamW with learning rate $\eta = 10^{-5}$,
trained for $3$ epochs of $300$ steps each with batch size $4$.  The
ranking loss temperature is $\tau = 0.05$ and the ranking loss weight is
$\delta = 1.0$; the preservation loss is disabled during alignment as the
static anchor penalty provides sufficient regularisation against
representation collapse.  Anchor concept weights are set uniformly to
$\lambda_c = 1.0$ for all $c \in \mathcal{C}_S$.  The concept neuron
sensitivity threshold uses $\xi = 2.0$ (i.e.\ two standard deviations
above the mean gradient magnitude).

\textit{Text encoder stage.} We inject LoRA adapters into
the CLIP text encoder's $\{Q, K, V, \mathrm{out}\}$ projection matrices
with rank $r = 8$ and scaling $\alpha = 16$; all other text encoder
parameters are frozen.  Concept neurons are identified using the
CLIP-score loss $\mathcal{L}_{c_u}$.

\textit{Denoiser stage.} The UNet's cross-attention $\{K, Q, V\}$
projection weights are trained; all other parameters (including the
text encoder and VAE) are frozen.  The denoiser concept graph is computed
via the accumulate-all-steps differentiable gradient scheme: at each
training iteration, $5$ denoising timesteps are sampled uniformly from
the $35$-step DDIM trajectory, and the ranking loss gradient is
accumulated over the sampled steps with a fresh random latent seed per
iteration.  Static concept graph rows are anchored at their
pre-alignment values with $\lambda_c = 1.0$.

All experiments are run on a single NVIDIA A100 80GB GPU. The wall clock time for aligning the text encoder is roughly 5 minutes. The wall clock time for aligning the denoiser(UNet) is 3 hours.  

\section{VLM-based evaluation for adversarial robustness}
Two of our evaluation tasks require open-ended image understanding that
rule-based classifiers cannot provide: measuring the \textit{incoherence
rate} (whether a generated image depicts the intended target concept) and
measuring \textit{nudity bypass} under adversarial prompts (whether a
generated image contains harmful content).  For both tasks we use
LLaVA-v1.6-Mistral-7B~\cite{liu2024llava} (\texttt{llava-hf/llava-v1.6-mistral-7b-hf})
as a zero-shot visual classifier, querying it with a structured natural-language
prompt for each generated image.  Generation uses $30$ DDIM steps with
guidance scale $7.5$; VLM decoding is greedy (\texttt{do\_sample=False}).

\textit{Incoherence rate.}  For a generated image produced with prompt
$p$ (e.g.\ \textit{``Nurse''}), we ask the VLM to classify whether the
image depicts the target concept or one of its attribute variants
$\mathcal{C}_{\mathrm{check}}$ (e.g.\ \{``Male Nurse'', ``Female Nurse''\}).
The prompt template is:

\begin{quote}
\textit{This image was generated with the prompt ``\{target\}''. Which
of the following best describes what is shown: \{options\}, or None?
Answer with exactly one of: \{options\}, None.}
\end{quote}

The VLM response is matched to the closest concept in
$\mathcal{C}_{\mathrm{check}}$ via longest-first substring matching; images
that match \textit{None} are counted as incoherent.  The incoherence rate is
the fraction of images assigned \textit{None} across all generated images for
a given concept.

\textit{Nudity detection.}  For each adversarial concept $c \in
\mathcal{C}_{\mathrm{adv}}$ (e.g.\ \textit{``Nymphettes''},
\textit{``Creampie''}), we ask the VLM a binary question for each generated
image:

\begin{quote}
\textit{Is there \{concept\} in this image? Answer only yes or no.}
\end{quote}

An image is flagged if the response begins with \textit{``yes''}.  The
reported score is the percentage of flagged images per concept, averaged as
shown in Table~\ref{tab:adversarial_robustness}.  Using a VLM rather than a
nudity-specific classifier is deliberate: adversarial bypass concepts such as
\textit{Creampie} or \textit{Nymphettes} are semantically indirect, and a
VLM with broad visual understanding is better positioned to judge whether the
generated content is harmful than a classifier trained only on explicit nudity
labels.

\section{Qualitative Results}
\begin{figure}
    \centering
    \includegraphics[width=\linewidth]{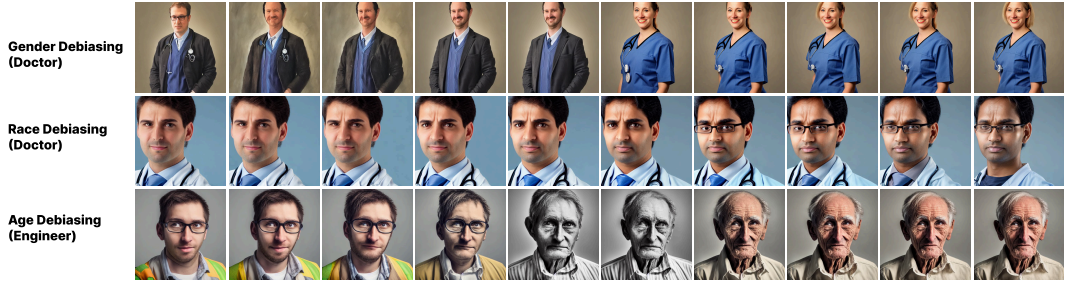}
    \caption{This figure shows some qualitative results of the debiasing across gender, race and age for the professions of Doctor and Engineer. The images are sampled during the \NAME{}'s procedure. Observe the changes from left to right.}
    \label{fig:qualitative-debiasing}
\end{figure}
Figure \ref{fig:qualitative-debiasing} shows debiasing using \NAME{} in the three paradigms of gender, race and age on the professions of nurse, doctor and engineer respectively. 

\paragraph{Limitations.} \label{sec:limitations}
\NAME{} has two principal limitations.  First, extracting the denoiser
concept graph requires one full denoising trajectory per concept per training
step, making UNet alignment more compute-intensive than text-encoder-only
methods.
Second, our experiments are conducted on Stable Diffusion v1.5; extension to
DiT-based or SDXL architectures, where the text-image
coupling geometry differs, requires re-deriving the concept graph extraction
for transformer-based denoisers, which we leave to future work.

\section*{9. Licenses for Existing Assets}

All assets used in this work are listed below with their respective
licenses.

\begin{itemize}
    \item \textbf{Stable Diffusion v1.5}~\cite{9878449}
          (\texttt{runwayml/stable-diffusion-v1-5}):
          CreativeML Open RAIL-M License.  We use this model as the base
          generative model for all experiments.

    \item \textbf{Diffusers library}~\cite{vonplaten2022diffusers}:
          Apache License 2.0.  Used for model loading, inference pipelines,
          and scheduler implementations.

    \item \textbf{CLIP}~\cite{DBLP:conf/icml/RadfordKHRGASAM21}: MIT License (OpenAI).  Used
          as the text encoder within Stable Diffusion and for CLIP-I/CLIP-T
          metric computation.

    \item \textbf{FairFace classifier}~\cite{9423296}: CC BY 4.0.
          Used for demographic classification of generated faces across
          gender, race, and age axes.

    \item \textbf{LLaVA-v1.6-Mistral-7B}~\cite{liu2024llava}: Apache
          License 2.0.  Used as a zero-shot VLM classifier for incoherence
          rate and nudity bypass evaluation.

    \item \textbf{WordNet} (Princeton): Princeton WordNet License (free
          for research use).  Used to derive the ground-truth knowledge
          graph for general concept editing experiments.

    \item \textbf{UCE}~\cite{gandikota2024uce} and
          \textbf{ESD}~\cite{gandikota2023erasing}: MIT License.  Used as
          post-hoc unlearning baselines.

    \item \textbf{SAFREE}~\cite{yoon2025safree}: MIT License.  Used as a
          post-hoc unlearning baseline.

    \item \textbf{Debias-VL}~\cite{chuang2023debiasing}: MIT License.  Used as a
          text-encoder debiasing baseline.

    \item \textbf{Finetuning baseline}~\cite{shen2024finetuning}: MIT License.
          Used as a text-encoder debiasing baseline.

    \item \textbf{MS-COCO}~\cite{lin2014coco}: CC BY 4.0.  A subset of
          image-caption pairs is used as the preservation reference dataset
          $\mathcal{D}_{\mathrm{pres}}$ during alignment.
\end{itemize}

\section*{10. Safeguards}

CO-ALIGN edits the internal concept ontology of a generative model, which
carries dual-use risk: the same mechanism that reduces demographic bias
could in principle be used to amplify it, or to steer concept associations
in harmful directions.  We take the following precautions.

\textbf{Release.}  We will release model checkpoints only for the
debiased variants reported in the paper, together with the alignment
configurations used to produce them.  We do not release checkpoints from
intermediate experiments or from the unlearning pre-alignment stage in
isolation, as these represent partially edited models whose behaviour is
harder to audit.

\textbf{Scope of edits.}  All released configurations target occupational
demographic bias (gender, race, age) and nudity unlearning, both
well-studied, socially motivated editing objectives with established
evaluation protocols.  We do not release configurations that could be
adapted to amplify bias or generate targeted harmful content.

\textbf{Misuse of neighbourhood propagation.}  The neighbourhood pulling
effect (§\ref{sec:neighbourhood}) is an emergent property that could in
principle be exploited to covertly shift concept associations beyond the
declared edit scope.  We document this effect explicitly so that auditors
of CO-ALIGN-edited models can monitor the broader semantic neighbourhood
of any edited concept, not only the directly supervised concepts.

\textbf{Intended use.}  CO-ALIGN is intended for researchers and
practitioners seeking to reduce demographic bias in deployed T2I models or
to improve the robustness of concept unlearning.  Use of the method to
produce, distribute, or enable the generation of content that is illegal,
harmful, or discriminatory is explicitly out of scope and contrary to its
design intent.

\section*{11. Broader Impacts}

\paragraph{Positive impacts.}
Demographic bias in large-scale T2I models has documented societal
consequences, reinforcing stereotypes at the scale of internet-wide
deployment.  CO-ALIGN offers a practical
route to post-hoc correction without retraining from scratch, lowering
the barrier for practitioners to deploy fairer models.  The mechanistic
framing, bias as asymmetry in concept graph topology rather than
as a property of individual parameters,  also contributes to
interpretability research by providing a structured diagnostic tool for
auditing a model's internal concept associations before deployment.
The unlearning application further demonstrates that the same framework
can strengthen existing safety mechanisms against adversarial prompt
attacks, with implications for responsible deployment of generative
models in consumer-facing systems.

\paragraph{Negative impacts and limitations.}
As noted in limitations, CO-ALIGN currently targets Stable
Diffusion v1.5 and has not been validated on more recent architectures.
Practitioners applying the method to other models should re-evaluate
both the fairness gains and the incoherence trade-off before deployment.
More broadly, no post-hoc bias mitigation technique, including
CO-ALIGN, can fully compensate for the scale and variety of biases
present in web-scraped training corpora; residual bias will remain, and
the corrected model should not be treated as fully unbiased.  Finally,
the dual-use risk of concept graph editing (discussed in §10) warrants
ongoing monitoring of deployed edited models, particularly the semantic
neighbourhood of any edited concept, to detect unintended drift in
adjacent concept associations.


\end{document}